\documentclass[conference]{IEEEtran}
\IEEEoverridecommandlockouts
\usepackage{cite}
\usepackage{amsmath,amssymb,amsfonts}
\usepackage{algorithmic}
\usepackage{algorithm}
\usepackage{graphicx}
\usepackage{textcomp}
\usepackage{xcolor}
\def\BibTeX{{\rm B\kern-.05em{\sc i\kern-.025em b}\kern-.08em
    T\kern-.1667em\lower.7ex\hbox{E}\kern-.125emX}}
\begin{document}

\title{A General Approach to Domain \\ Adaptation with Applications in Astronomy}

\author{\IEEEauthorblockN{Ricardo Vilalta$^*$, Kinjal Dhar Gupta, Dainis Boumber}
\IEEEauthorblockA{\textit{Department of Computer Science} \\
\textit{University of Houston}\\
Houston TX, 77204-3010, USA \\
$^*$corresponding author email: rvilalta@uh.edu}
\and
\IEEEauthorblockN{Mikhail M. Meskhi}
\IEEEauthorblockA{\textit{Department of Computer Science} \\
\textit{North American University}\\
Stafford TX, 77477, USA}}

\maketitle

\begin{abstract}
The ability to build a model on a source task and subsequently adapt such model on a new target task is a pervasive need in many astronomical applications. The problem is generally known as \textit{transfer learning} in machine learning, where \textit{domain adaptation} is a popular scenario. An example is to  build a predictive model on spectroscopic data to identify Supernovae IA, while subsequently trying to adapt such model on photometric data. In this paper we propose a new general approach to domain adaptation that does not rely on the proximity of source and target distributions. Instead we simply assume a strong similarity in model complexity across domains, and use active learning to mitigate the dependency on source examples. Our work leads to a new formulation for the likelihood as a function of empirical error using a theoretical learning bound; the result is a novel mapping from generalization error to a likelihood estimation. Results using two real astronomical problems, Supernova Ia classification and identification of Mars landforms, show two main advantages with our approach: increased accuracy performance and substantial savings in computational cost.
\end{abstract}


%% main text
\section{Introduction}
%***********************************
\label{sc:introduction}

% What are we doing to solve the problem?
In this paper we propose a new approach to domain adaptation that is particularly well suited for astronomical applications. Our general setting assumes the learner is embedded in a \textit{domain adaptation} framework
\cite{BenDavid06,BenDavid10,Blitzer06,Blitzer07,Daume06,Mansour09,Daume09,Bruzzone10}, where the goal is to obtain a predictive model on a target domain, where examples abound, but labeled data is scarce. We assume the existence of a source domain with abundant labeled data, but with different distribution, such that the naive approach of directly applying the source model on the target becomes inadequate. Instead we follow a Maximum A Posteriori (MAP) approach to estimate model complexity by extracting the prior distribution from previous experience (i.e., from a previous task), and by taking the (scarce) target data as evidence to compute the likelihood. The result is a new approach to domain adaptation that is exempt from the common restriction that demands close proximity between source and target distributions \cite{BenDavid10}.

% Advantages with our approach
We show how using a prior distribution from a previous task to estimate a posterior of model complexity on a new task, not only yields an increase in accuracy performance, but in addition has an enormous impact on computational cost. Our focus is on astronomical problems where we are witnessing a rapid growth of data volumes corresponding to a variety of astronomical surveys; data repositories have gone from gigabytes into terabytes, and we expect those repositories to reach the petabytes in the coming years \cite{Brescia13,Feigelson17}. Our proposed methodology assumes an exhaustive search for the right model complexity on a source domain, where we generate a prior distribution on model complexity. The arrival of a new target task dispels with such exhaustive search; instead it generates a posterior distribution that directly leads to finding a near-optimal figure of model complexity. This is particularly important for big-data applications where lengthy computational tasks are unavoidable, even under the availability of an efficient high-performance-computing infrastructure.

% Experiments
We report on experiments using two real-world astronomical domains: classification of Supernovae Ia using photometric data, and characterization of landforms on Mars using Digital Elevation Maps (DEMs). Both domains can produce massive amounts of data with a strong need for efficient computational solutions. Results show how the use of a source prior to guide the search for an optimal value of model complexity can significantly improve on generalization performance.

% Comments
The rationale for assuming similar model-complexity values across tasks is based on the nature of distributional discrepancies in many physical domains. The idea is useful not only to astronomical data analysis, but to many other real-world problems where the shift in distribution originates from more sophisticated equipment (e.g., modern telescopes), different instrumentation, or different coverage on the feature space, while the complexity of the classification problem experiences little change. For example, while spectroscopic and photometric observations capture data at different levels of resolution, the identification itself of specific astronomical objects shares a similar degree of difficulty.
In short, we assume that the change in distribution from source to target does not affect model complexity significantly.

% Organization
This paper is organized as follows. We begin by providing basic concepts in classification and domain adaptation, followed by a detailed description of our proposed approach that shows how to extract a prior distribution from a source domain. We then show our experiments and empirical results. The last section provides summary and conclusions.

%=============================================================================

\section{Preliminary Concepts}
%***********************************
\label{sc:preliminaries}

\subsection{Basic Notation}
%---------------------------
\label{ssc:notation}

In supervised learning or classification, we assume the existence of a training set of examples, $T = \{({\bf x_i}, y_i)\}_{i=1}^{p}$, where vector ${\bf x} = (x_1, x_2, \cdots, x_n)$ is an instance of the input space ${\cal X}$, and $y$ is an instance of the output space ${\cal Y}$. It is often assumed that sample $T$ contains independently and identically distributed (i.i.d.) examples that come from a fixed but unknown joint probability distribution, $P(\mathbf{x},y)$, in the input-output space ${\cal X} \times {\cal Y}$. The output of the learning algorithm is a function $f_{\theta}(\mathbf{x})$ (parameterized by $\theta$) mapping the input space to the output space, $f_{\theta}: \cal{X} \rightarrow {\cal Y}$. Function $f_{\theta}$ comes from a space of functions $\mathcal{H}$. The idea is to search for the hypothesis that minimizes the expectation of a loss function $L(y,f(\mathbf{x}|\theta))$, a.k.a. the risk:

\begin{equation}
R(\theta,P(\mathbf{x},y)) = E_{\sim P}[L(y,f(\mathbf{x}|\theta))]
\end{equation}

\noindent where we usually employ the zero-one loss function:

\begin{equation}
L(y,f(\mathbf{x}|\theta)) = 1_{\{ \mathbf{x} | \mathbf{y(x)} \neq f(\mathbf{x}|\theta)\}} (\mathbf{x})
\end{equation}

\noindent
such that $1(\cdot)$ is an indicator function, and $\mathbf{y(x)}$ is the true class of $\mathbf{x}$.

\vspace*{2mm}
\noindent
\textbf{Domain Adaptation}.
In domain adaptation, we assume the existence of a source domain, corresponding to a previous task from which experience can be leveraged, and a target domain, corresponding to the present task. Each domain enables us to draw a dataset: $T_s = \{({\bf x_i}, y_i)\}_{i=1}^{p}$ for the source, and $T_t = \{{\bf x_i}\}_{i=1}^{q}$ for the target. $T_s$ is an instantiation of a joint probability distribution, $P_s(\mathbf{x},y)$, while $T_t$ is an instantiation of the marginal distribution $P_t(\mathbf{x})$ (from the joint distribution $P_t(\mathbf{x},y)$, such that $P_t(\mathbf{x}) = \int_y P_t(\mathbf{x},y) d_y$). The emphasis is always placed on the target domain, corresponding to the task at hand. The main objective is to induce a model from the target dataset; when building the model, one can exploit knowledge from the source dataset.  A major difficulty in domain adaptation stems from the lack of labels on $T_{\rm t}$. We will assume the possibility of \textit{querying} some of those examples to attain a few labeled examples as part of an active learning setting \cite{settles2012}.

Most domain adaptation methods assume similar class posteriors across source and target domains, i.e., $P_{\rm s}(y | \mathbf{x}) = P_{\rm t}(y | \mathbf{x})$, but different marginals $P_{\rm s}(\mathbf{x}) \neq P_{\rm t}(\mathbf{x})$; this is known as the covariate-shift assumption. Different from previous work, we will consider the case where both source and target differ in the marginal distributions and class posteriors.

\vspace*{2mm}
\noindent
\textbf{Parameter Estimation}.
We also consider the problem of parameter estimation, which can play a major role in classification as a means to estimate an optimal figure of model complexity. Examples include finding the number of hidden nodes in a neural network, or finding the degree of a polynomial kernel in support vector machines. In Maximum a Posteriori (MAP), the goal is to obtain a point estimate that maximizes the posterior distribution of the parameter given the data or evidence. The posterior probability is essentially a function of two main factors: the prior probability (i.e., degree of belief of model complexity before data analysis) and the likelihood (i.e., probability of data sample conditioned on model complexity). When data abounds, the likelihood bears a stronger influence on the posterior, while the opposite takes place when data is scarce; here the prior bears more influence on the posterior. An important question is how to obtain a reliable prior when data is scarce (i.e., when the prior plays a strong role in estimating the posterior).

\subsection{Related Work}
%---------------------------
\label{sc:related_work}

Domain adaptation induces a model by exploiting experience gathered from previous tasks \cite{BenDavid10}. It is considered a subfield of transfer learning \cite{pan2010survey}, and has become increasingly popular in recent years due to the pervasive nature of task domains exhibiting differences in sample distribution \cite{Liu15,Xu14}. The central question is if a previously constructed (source) model can be adapted to a new task, or if it is better to build a new (target) model from scratch.

Domain adaptation papers can be classified into two types: instance-based and feature-based methods. The idea in instance-based methods is to assign high weights to source examples occupying regions of high density in the target domain. A popular approach is known as covariate shift \cite{Quinonero09,Shimodaira00,Kanamori09,Sugiyama09,Bickel09}. The covariance-shift assumption is that one can build a model on the newly-weighted source sample and apply it directly to the target domain \cite{sugiyama2008direct,gretton2009covariate}. A stringent requirement is that source and target distributions must be close to each other.

Feature-based domain adaptation methods attempt to project source and target datasets into a latent feature space, where the covariate-shift assumption holds. A model is then built on the transformed space, and used as the classifier on the target. Examples are structural corresponding learning \cite{Blitzer06}, subspace alignment methods \cite{fernando2013unsupervised}, among others \cite{ando2005framework,glorot2011domain,Bruzzone10}.

From a theoretical view, previous work has tried to estimate the distance between source and target distributions \cite{BenDavid06,BenDavid10,Blitzer07}; and employ regularization terms to find models with good generalization performance on both source and target domains \cite{Daume10}.

%=============================================================================

\section{Methodology}
%*******************************************
\label{sc:methodology}

We begin by providing a general description of the proposed methodology. The main idea is to assist in finding the right configuration (model complexity) for a learning algorithm by leveraging information from a previous similar task. For example, when trying to find a predictive model to classify Supernovae, or to predict the class of a transient star, searching for a model with the right degree of complexity by varying a configuration parameter(s) may turn frustratingly cumbersome. As an example, setting the architecture of a (shallow) neural network by varying the number of hidden nodes would lead to a huge number of experiments to assess model quality for each architecture. To alleviate this situation, we \textit{learn} a range of values of model complexity from a previous task using a Maximum a Posteriori approach, where there is a high likelihood of finding a good value of model complexity (e.g., number of hidden nodes) on the new task. Moreover, different from previous work, our method disregards many assumptions made by previous work: we do not follow the covariate-shift assumption; no data projection is required to transform the feature space (thus incurring no loss of information); and the dependence on the source is not based on transferring source examples to build the target model. To summarize, the main idea is to learn about the model-building process employed in a previous task (source domain), and to transfer that experience to the new task (target domain). Experience is here understood as a distribution of optimal values of model complexity.

\subsection{Active Learning}
%---------------------------------------------------------------
\label{ssc:active-learning}

Our basic strategy is to step aside from the common approach followed by many domain-adaptation techniques that selectively gather source examples to enlarge the set of target examples. When source and target distributions differ significantly, such approach can lead to a biased model. Under high distribution discrepancy, an optimal strategy would simply rely on target instances. But the classical setting in domain adaptation provides none (or very few) class labels on the target dataset. A solution to this conundrum is to provide target class labels using active learning \cite{settles2012,Balcan09,Cohn96}, where a selective mechanism queries an expert for (target) class labels under a limited budget (i.e., a limited number of queries).

The use of active learning in domain adaptation obviates using source examples while building the target model, opening new research avenues in the field of transfer learning. Here we investigate a mechanism that generates a distribution of model complexity on the source domain, and re-utilizes such distribution as a prior in a Bayesian setting over the target domain. The resulting point-estimate over the posterior distribution of model complexity depends on the prior (source domain) and the likelihood, or evidence (target domain).

\subsection{Model Complexity as a Transferable Item}
%---------------------------------------------------------------
\label{ssc:transferable}

We assume that optimal predictive models for both source and target domains share a similar degree of model complexity. For example, assuming both domains are best modeled using Support Vector Machines with a polynomial kernel parameterized by $\theta$ (corresponding to the degree of a polynomial), we can then further assume that $P_s(\theta^*) \sim P_t(\theta^*)$, where $\theta^*$ is the polynomial degree that minimizes a loss function. Such assumption focuses on the similarity of complexity-parameter distributions, and not on the similarity of joint input-output distributions.

Specifically, iteratively sampling and building predictive models on the source domain leads to a distribution of model parameters, $P_s(\theta)$. Our goal is then to estimate an optimal parameter value $\theta^*$ that maximizes the posterior distribution on the target domain $P_t(\theta|D)$, where $D$ is the data or evidence (i.e., target sample $T_t$). By using the distribution gathered from the source domain as a reliable prior, we can formulate the posterior using Bayes formula:

\begin{equation}
P_t(\theta|D)=\frac{P_{t}(D|\theta)P_{s}(\theta)}{\sum_iP_{t}(D|\theta_i)P_{s}(\theta_i)}
\label{eq:posterior}
\end{equation}

\noindent
where $P_{t}(D|\theta)$ is the likelihood, and $P_{s}(\theta)$ the prior. This is precisely how we propose to adapt a model across domains; assuming the complexity of the  model built on the source domain is similar to that on the target domain, we look at the source prior $P_s(\theta)$ as the \textit{transferable item} to be used in the new target domain.

Since the denominator in Eq.~(\ref{eq:posterior}) is constant, we can simplify the equation as follows:

\begin{equation}
\label{eq:prop_posterior}
P_{t}(\theta|D) = Z P_{t}(D|\theta) P_{s}(\theta) \propto P_{t}(D|\theta) P_{s}(\theta)
\end{equation}

\noindent
where $Z$ is a normalization factor. To optimize $P_{t}(\theta|D)$, we optimize the product of $P_{t}(D|\theta)$ and $P_{s}(\theta)$, and disregard the value of $Z$, as it is not a function of parameter $\theta$. Hence, our goal is not to obtain a distribution for the posterior $P_{t}(\theta|D)$, but only to estimate the value of $\theta$ that maximizes the product of the likelihood and prior\footnote{This is different from Bayesian estimation where the output is a full posterior probability distribution over model parameters \cite{Bolstad07,Goodman05,Louis05}.}, a technique known as Maximum a Posteriori, a.k.a. MAP.
We now explain how to compute the prior $P_s(\theta)$ and likelihood $P_t(D|\theta)$ to obtain a point estimate of model complexity on the target domain.

%=============================================================

\subsection{Estimating the Likelihood}
%---------------------------------------------------------------------------------
\label{ssc:likelihood_estimate}

Our approach to estimate the likelihood is as follows. We first use active learning to obtain a (small) labeled sample from the target domain. We then introduce a novel mechanism to compute $P_t(D|\theta)$ by mapping generalization error to a likelihood probability. We explain both steps next.

\vspace*{2mm}
\subsubsection{Active Learning}
%________________________________
\label{sssc:active_learing}

To lessen the dependence on the source domain, we resort to active learning to produce an informative sample of labeled instances from the target domain. We use pool-based active learning with margin sampling \cite{scheffer2001active} as the uncertainty sampling  technique \cite{lewis1994sequential,lewis1994heterogeneous}. Specifically, the algorithm randomly selects an initial set of instances from the unlabeled target dataset $T_{t}$, and queries their class labels; it then iteratively builds a model $f_t(\mathbf{x}|\theta)$ on the labeled target instances as follows. At every iteration, the algorithm identifies the instance $\mathbf{x}_i$ from the remaining unlabeled target instances with the minimum margin (i.e., minimum distance to the decision boundary),  queries $\mathbf{x}_i$ to obtain class label $y_i$, and adds $(\mathbf{x}_i,y_i)$ to the set of labeled target instances. The process repeats until a budget (i.e., maximum number of allowed queries) is exhausted. The result is a labeled sample that will be used to compute the likelihood $P(D|\theta)$.

\vspace*{2mm}
\subsubsection{Mapping Error to a Likelihood}
%_________________________________________________
\label{sssc:new_estimate}

In general, the likelihood $P(D|\theta)$ estimates the probability of seeing data $D$ given parameter $\theta$. This estimation is particularly complex in our study because $\theta$ is normally understood as a parameter of a probabilistic or generative model. $P(D|\theta)$ indicates how likely it is to obtain $D$ from a probabilistic model parameterized by $\theta$. In our case $\theta$ is unconventionally defined as a (complexity) parameter of a predictive model $f(\textbf{x}|\theta)$ (e.g., degree of a polynomial kernel).

\vspace*{0mm}
\noindent \hrulefill
\vspace*{0mm}

\noindent We contend that $P(D|\theta)$ can be re-interpreted as \textit{the probability of $D$ given the
empirical error of $f(\textbf{x}|\theta)$} on $D$. In general, the lower the empirical error,
the higher the likelihood that model $f(\cdot)$ can reproduce the class labels contained in $D$.

\vspace*{-2mm}
\noindent \hrulefill

\vspace*{2mm}
Our definition of empirical error on $D$ refers to the error incurred on sample $D$ alone, and
is denoted as a function of $\theta$ and $D$, $\widehat{g}(\theta|D)$. Empirical error is also
known as in-sample error.

Our re-interpretation of the likelihood leads naturally to the assumption that the probability of $D$ being classified correctly by a hypothesis $f(\mathbf{x}|\theta)$ is inversely proportional to the error
made by $f(\cdot)$ on $D$. We formulate this inverse relation assuming an exponential
distribution:

\begin{equation}
P(D|\theta) = \lambda \exp({-\lambda \widehat{g}(\theta|D)})
\label{eq:likelihood}
\end{equation}

\noindent
where $\lambda$ is the rate parameter. This formulation simply states that the likelihood $P(D|\theta)$
decreases exponentially with error $\widehat{g}(\theta|D)$, but it is clearly handicapped, as
different values of complexity $\theta$ are mapped to the same likelihood as long as the empirical error
$\widehat{g}(\theta|D)$ is identical. However, under equal error rates, we would like to
assign a higher likelihood to simpler models. We propose a solution to this next.

\vspace*{3mm}
\begin{figure}[htb]
\begin{center}
\includegraphics[scale=0.25]{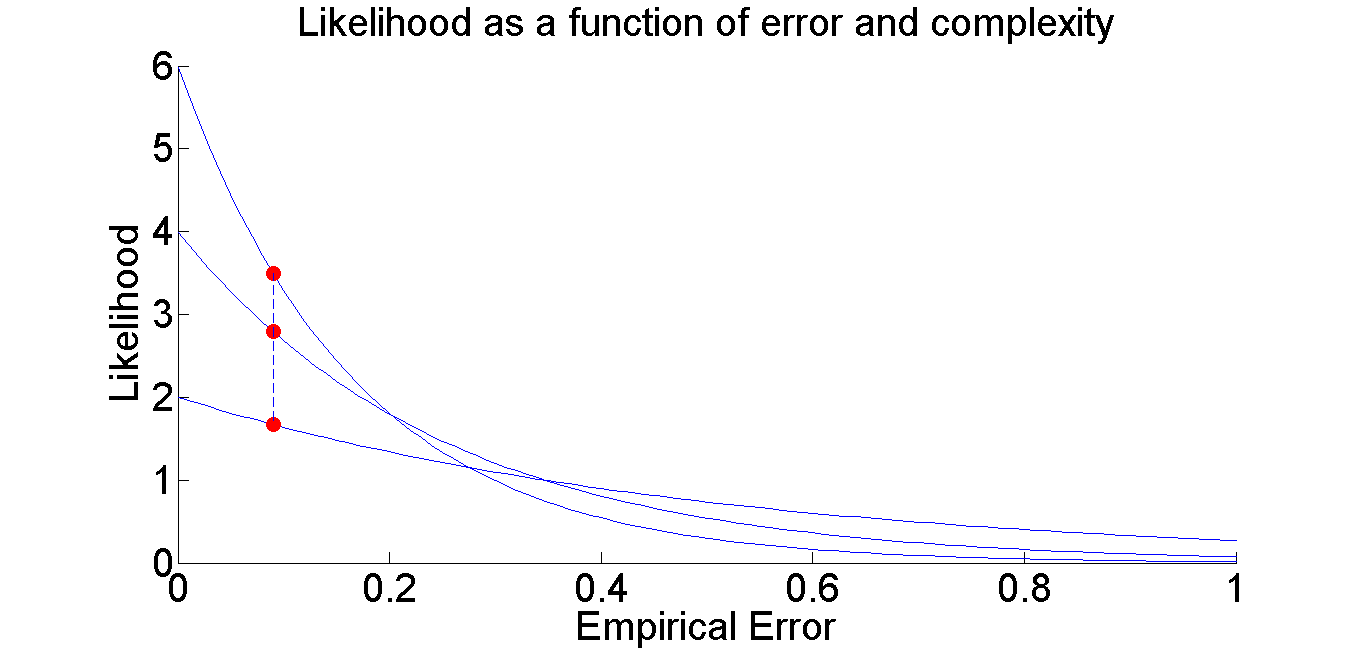}
\caption{Likelihood of the data given 1) empirical error $\widehat{g}(\theta|D)$ and 2) a scaled version of error variance as a function of the VC-dimension $d_{\mathrm{VC}}(\mathcal{H})$. Equal values for $\widehat{g}(\theta|D)$ do not map into the same likelihood.}
\label{fig:likelihood}
\end{center}
\end{figure}

\subsubsection{Adding Robustness to the New Likelihood}
%________________________________________________________
\label{sssc:robust_estimate}

Our formulation of the likelihood as a function of empirical error (Eq.~(\ref{eq:likelihood})) can
be made more robust by taking into account the variance component of error induced by models that
belong to families exhibiting high VC-dimension (Vapnik-Chervonenkis dimension \cite{Haussler91}). In short, we suggest to penalize those scenarios where VC-dimension is high. To start, we define $g(\theta|D)$ as the expected error across the whole input-output distribution:

\begin{equation}
g(\theta|D) = \int_\mathcal{X} \int_\mathcal{Y}  L(y,f(\mathbf{x}|\theta)) P(\mathbf{x},y) dxdy
\label{eq:error}
\end{equation}

\noindent
where the loss $L(y,f(\mathbf{x}|\theta)$ is the zero-one loss function.
$g(\theta|D)$ is also known as generalization error. Now, we know from
Vapnik-Chervonenkis inequality \cite{abu2012learning} that with
probability $1-\delta$, an upper bound on $g(\theta|D)$ is given as follows:

\begin{equation}
\label{eq:error_bound}
g(\theta|D) \leq \widehat{g}(\theta|D) + \sqrt{\frac{8}{N}\ln\frac{4m_H(2N)}{\delta}}
\end{equation}

\noindent
where $\delta$ is user defined, $N$ is the number of training instances, and $m_\mathcal{H}(q)$ is a polynomial function that defines that largest number of dichotomies on $q$ training instances given the class of hypotheses $\mathcal{H}$:

\begin{equation}
\label{eq:NumberOfDichotomies}
m_H(q) \leq \sum_{i=0}^{d_{\mathrm{VC}}(H)} {\frac{N!}{i! N-i!}}
\end{equation}

\noindent
where $d_{\mathrm{VC}}(H)$ is the VC-dimension \cite{Haussler91}, defined as the maximum number of examples that can be shattered by $\mathcal{H}$ \cite{abu2012learning}, and depends on the complexity of the hypothesis (i.e., on $\theta$). The VC-dimension of various classes of hypotheses is well known. For example, the VC-dimension $d_{\mathrm{VC}}(\mathcal{H})$ of neural networks with a sigmoid gate function has a lower bound of $\sigma(w \log w)$ and an upper bound of $O(w^2)$ \cite{maass1995vapnik}, where $w$ is the number of weights in the network. In this example, parameter $\theta$ can be interpreted as the number of hidden nodes $h$ in a feed-forward neural network (NN). The $d_{\mathrm{VC}}(H)$ of a neural network (NN) with $h$ hidden nodes can then be estimated using the lower bound $w \log w$ and defining $w$=($i$ +1)$\times$ $h$ + ($h$ + 1)$\times$ $o$, where $i$ and $o$ are the number of input features and output classes respectively.

We now show how to strengthen our definition of the likelihood. In essence, we keep the exponential distribution the same (Eq.~(\ref{eq:likelihood})), but make parameter $\lambda$ a function of model complexity $\theta$, $\lambda(\theta)$:

\begin{equation}
\label{eq:new_full_likelihood}
P_t(D|\theta) = \lambda(\theta) \exp({-\lambda(\theta) \widehat{g}(\theta|D)})
\end{equation}

\noindent
and define function $\lambda(\theta)$ as a scaled version of the second part of the Vapnik-Chervonenkis
inequality (Eq.~(\ref{eq:error_bound})):

\begin{equation}
\label{eq:lambda}
\lambda(\theta) = \alpha \sqrt{\frac{8}{N}\ln\frac{4m_\theta(2N)}{\delta}}
\end{equation}

\noindent
where $\alpha$ is a user-defined scale factor that decides how much weight is placed on the variance component of error. By transforming parameter $\lambda$ into a function parameterized by $\theta$, $\lambda(\theta)$, we achieve our goal of assigning higher likelihoods to simpler models when comparing hypotheses showing similar empirical error.  We illustrate these concepts in Figure~\ref{fig:likelihood}.

The ideas mentioned above have been tried using different strategies. Examples include a full Bayesian approach in transfer learning that finds a common subspace across tasks using a kernel-based dimensionality-reduction technique \cite{Gonen14}; transferring priors across multiple tasks using a Hierarchical Bayesian approach \cite{Finkel09}; finding clusters of tasks under a Dirichlet process prior \cite{Roy07}; finding a Gaussian prior from previous tasks \cite{Raina06}; and theoretical studies using the PAC learning framework on a Bayesian setting \cite{Germain16}. All these methods are contingent on the proximity of source and target distributions, whereas our approach relies primarily on the similarity of model complexity.

Embedding the empirical error in an exponential function to compute the likelihood has been tried before \cite{Germain16B}, albeit without considering the capacity of each hypothesis. The novelty of our approach lies in transforming the likelihood function according to the VC-dimension of $\mathcal{H}$.

%=============================================================

\subsection{Estimating The Prior}
%---------------------------------------------------------------------
\label{ssc:prior_estimate}

Regarding the prior distribution of $\theta$  (model complexity) on the source domain, we adopt a parametric model assuming a univariate Gaussian distribution, $$P_s(\theta)=\frac{1}{\sigma \sqrt{2\pi}} e^{-\frac{1}{2} (\frac{\theta-\mu}{\sigma})^2}$$ where $\mu$ and $\sigma^2$ are the mean and variance respectively. Specifically, our methodology generates $k$ samples of the source dataset $T_s$ using uniform random sampling without replacement; $k$ is user-defined, and can be regarded as an experimental design parameter.

We construct classifiers on each of the $k$ samples using a range of model complexity values, $\theta$ $\in$ $\lbrace\theta_1, \theta_2, \theta_3,.. \theta_m\rbrace$. For each sample $S_i$,  we find a value $\theta_i^*$, $1 \leq i \leq m$, that minimizes the expected loss (i.e., maximizes accuracy). The result is a sample with $k$ optimal values of model complexity. Our estimate of the prior is finally obtained by fitting these values to the Gaussian model.

%=============================================================

\subsection{Estimating the Posterior}
%----------------------------------------------------
\label{Posterior_Estimate}

Once we have estimated the prior $P_s(\theta)$  and likelihood $P_{t}(D|\theta)$, we can estimate
the numerator of the posterior distribution (Eq.~(\ref{eq:prop_posterior})): $P_{t}(\theta|D) = Z P_{t}(D|\theta) P_{s}(\theta) \propto P_{t}(D|\theta) P_{s}(\theta)$. Since we are interested in obtaining a point estimate for the posterior, we look for an optimal value $\theta^* = \arg\max_{\theta} P_t(D|\theta)P_s(\theta)$.

To reduce the space of complexity values during optimization, we limit the values of $\theta$ to the range $[\mu - \sigma, \mu + \sigma]$, where $\mu$ and $\sigma$ are the mean and standard deviation of the source prior distribution $P_s(\theta)$. The final value $\theta^*$ is used to build a classifier $f_t(\mathbf{x}|\theta^*)$ on the target domain using the queried instances as our training dataset.

Our methodology is outlined in Algorithm~\ref{algo:da_al}. The algorithm assumes as input the labeled source dataset $T_s$, the unlabeled target dataset $T_t$, the size $r$ of a small labeled sample to generate a model on the target, and a budget $b$ of possible queries to obtain additional labeled instances on the target. The first step is to build a prior distribution $P_s(\theta) \sim N(\mu, \sigma)$ on the source domain by exhaustively looking for an optimal figure of model complexity; this step can be computationally expensive, but can save substantial amounts of time when it is re-used on a future (target) task. The next steps compute the likelihood $P_{t}(D|\theta_i)$ in an iterative manner, by using labeled instances from the target obtained through active learning. The search is made narrow by limiting values of model complexity to just one standard deviation away from the source mean. The last steps build a (proportional) posterior distribution, and a predictive model using our optimal point estimate for model complexity.

% --- Algorithm 1 ---------------------------------------------------------
\begin{algorithm} [!ht]
\caption{Model Complexity Estimation Using Domain Adaptation and Active Learning}
\label{algo:da_al}
\begin{algorithmic}
\STATE \textbf{Input :} Source Dataset $T_s$, Target Dataset $T_t$,
Budget $b$, Initial Sample Size $r$.
\STATE \textbf{Output :} Predictive Target Model $f_t^*(\textbf{x}|\theta)$
\end{algorithmic}
\begin{algorithmic}[1]
\STATE Estimate prior $P_s(\theta) \sim N(\mu, \sigma)$ using source dataset $T_s$
\STATE Set $\theta_{\min} = \mu - \sigma$ and $\theta_{\max} = \mu + \sigma$
\STATE Use the small set of $r$ labeled instances from $T_t$ to build model $f_t(\textbf{x}|\theta)$
\STATE Use $f_t(\textbf{x}|\theta)$ and active learning to label $b$ target instances from $T_t$
\FOR{$\theta_i \gets \theta_{\min}, \theta_{\max}$}
\STATE Build model $f_t^i(\textbf{x}|\theta)$ with $\theta_i$ as model complexity
\STATE Compute $\widehat{g}(\theta_i|D)$ and $\lambda(\theta_i)$ to estimate likelihood $P_{t}(D|\theta_i)$
\STATE Estimate (proportional) posterior: $P_{t}(D|\theta_i) P_s(\theta_i)$
\ENDFOR
\STATE Let $\theta^* = \arg\max_{\theta_i} P_t(D|\theta_i)P_s(\theta_i)$
\STATE Build $f_t(\textbf{x}|\theta^*)$
\STATE \textbf{return} $f_t(\textbf{x}|\theta^*)$
\end{algorithmic}
\end{algorithm}

%=============================================================================

\section{Experiments}
%*******************************************
\label{sc:experiments}

We describe our experiments in detail next. All our code and datasets have been made available for reproducibility as a github project\footnote{Please visit https://github.com/PAL-UH/transferAL}.

We report empirical results on two different scientific areas to validate our methodology. The first area refers to the automatic classification of supernovae using photometric light curves. The second area is centered on the classification of landforms on planet Mars using digital elevation maps.

\subsection{Supernova Datasets}
%-----------------------------------
\label{ssc:supernova}

The automatic identification of Supernovae Ia (SNe Ia) has become a key step in many astronomical endeavors
\cite{Blondin12,Sasdelli16}. Among different types of supernovae,  SNe Ia are of particular relevance because they can be used as standard candles to probe large cosmological distances. The classification goal here is to identify SNe Ia (positive class) among other types (SNe Ib and Ic, negative class).

When analyzing light from a supernova, one can either exploit spectroscopic measurements, to take advantage of the wealth of information that can be obtained from spectral data. Such approach, however, is laborious and cumbersome. Another more common approach is to exploit photometric measurements; these are easier to obtain, but limited to a summarization of light intensity in bands or filters.  The domain adaptation framework fits into this scenario as follows \cite{DharGupta16}: the source dataset corresponds to spectroscopic measurements where class labels (SNe Ia, Ib and Ic) are known with high confidence (but data is scarce); whereas the target domain corresponds to photometric measurements where class labels are missing (but data abounds).

Our experiments use simulations to generate samples that resemble the type of measurements expected when using spectroscopic or photometric measurements. This brings the advantage of having samples with the same set of features (i.e., same input space $\mathcal{X}$). Specifically, we use data from the Supernova Photometric Classification Challenge \cite{kessler2010supernova}, consisting of supernova light curves simulated according to Dark Energy Survey specifications, using SNANA light curve simulator \cite{Ishida13}. The data comes from simulations that approximate the characteristics of the Dark Energy Survey (DES). Simulations include both spectroscopic (source) and photometric (target) samples; these are created with biases found in true datasets, where spectroscopic data are in general smaller, brighter, closer, and less noisy than photometric data.

Regarding the construction of simulated samples, we follow the data processing steps specified in \cite{Ishida13}. We only take objects with a minimum of three observed epochs per filter; at least one of them occurs before -3 days and at least one after +24 days since maximum brightness. On each filter, we use Gaussian process regression to do light-curve model fitting \cite{Chilenski15}; the resulting function is sampled using a window of size one day. No quality cuts are imposed (SNR$>0$). At the end, the spectroscopic (source) sample has 718 SNe, while the photometric (target) sample has 11946 SNe. Both samples have 108 columns or features (27 epochs $\times$ 4 filters). We use Kernel Principal Component Analysis (KPCA) to reduce the original dataset from 108 features to solely 20 features. This form of dimensionality reduction is useful to avoid the curse of dimensionality \cite{Hastie01}. A preliminary analysis shows no loss of information or accuracy performance after data transformation using KPCA.

\subsection{Mars Landforms}
%---------------------------------
\label{ssc:mars-landforms}

The second area corresponds to the automatic geomorphic mapping of planetary surfaces.
Specifically, the goal is to label segments on specific regions on planet Mars with their
corresponding landforms \cite{Bue06,Ghosh06,Stepinski05}. The input data comes in the form of raster data or digital elevation models (DEMs) produced by orbiting satellites (Mars Orbiter Laser Altimeter instrument on board the Mars Global Surveyor spacecraft). Each DEM is first subdivided into meaningful segments (groups of adjacent pixels with similar terrain properties) that are subsequently classified into the following landforms: crater floors, convex crater walls, concave crater walls, convex ridges, concave ridges, and inter-crater plateau. Domain adaptation is important to attain accurate predictive models on new target sites that exhibit different distribution from the original source site.

\begin{figure}[tbh]
\vspace*{0mm}
\begin{center}
\includegraphics[scale=0.48]{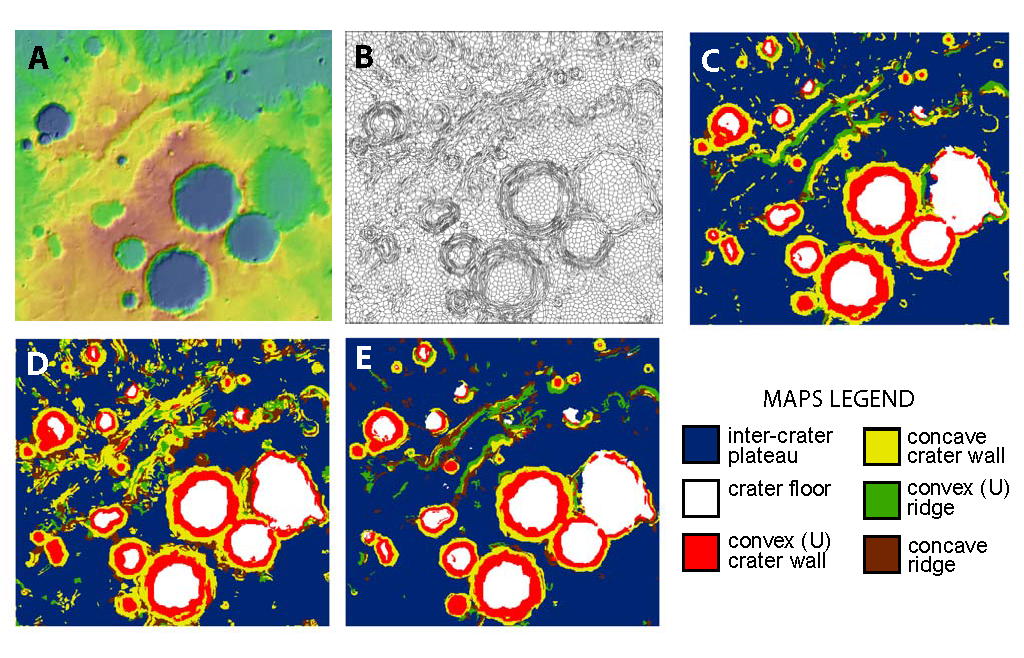}
\vspace*{-2mm}
\caption{DEMs on Mars are processed and classified into different landforms. The original DEM (A), corresponding to a site known as Tissia Valles, is segmented for labeling processes (red-to-blue gradient indicates high-to-low elevation). (B) The DEM is then classified using models of different complexity (C-E; color labels explained on bottom-right). This site, Tissia Valles, acts as the source domain. }
\label{fig:Landforms-Mars}
\end{center}
\vspace*{0mm}
\end{figure}

Figure~\ref{fig:Landforms-Mars} illustrates the sequence of steps needed to classify landforms on Mars using DEMs. The original map (A) is first divided into small segments amenable to labeling (B).  A model is then trained on a fraction of all segments and applied to the rest. Models of different complexity yield different classifications (C-E).

\begin{figure}[tbh]
\vspace*{0mm}
\begin{center}
\includegraphics[scale=0.32]{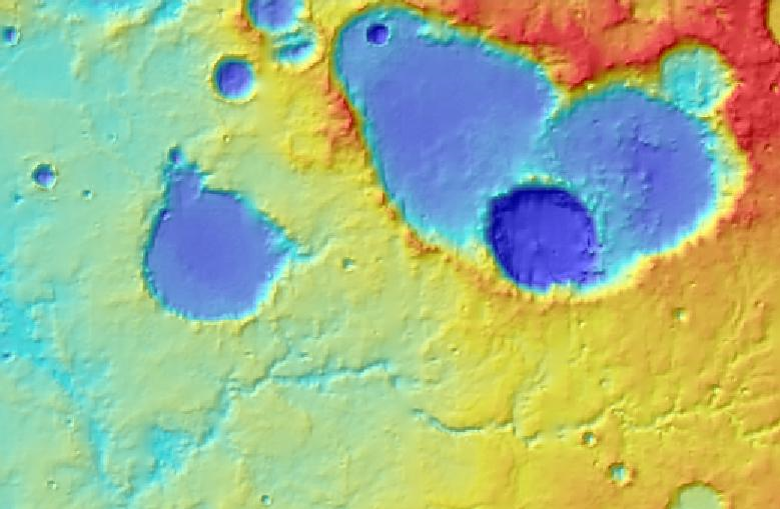}
\vspace*{-3mm}
\caption{A digital elevation map DEM of a region on Mars known as Evos, corresponding to our target domain (red-to-blue gradient indicates high-to-low elevation).}
\label{fig:Evos}
\end{center}
\vspace*{-2mm}
\end{figure}

The source domain is a region on Mars where we know the labels for all landforms; it is shown in  Figure~\ref{fig:Landforms-Mars}(A) and is known as Tissia Valles; it was chosen primarily because in a relatively small area, most landforms of interest are present. The region is heavily cratered and many different crater morphologies are present in a range of sizes. The goal here is to leverage experience during the model building process on Tissia Valles to find the right complexity for a model induced on a new site on Mars, corresponding to the target domain, where the landform distribution is different. In our experiments, the target site corresponds to a region known as Evos, shown in Figure~\ref{fig:Evos}. Notice the difference in distribution, where the shape of the craters and number of them differs significantly from Tissia Valles.

The input to the learning algorithm is not the DEM, but a training set made of feature vectors, one for each segment in the map. A segment is made of adjacent pixels with similar feature values. Each segments is characterized by three features, which are averages over the pixels embedded by the segment: slope, computed as the maximum rate of change in elevation from a cell to its neighbors (indicative of the steepness of the terrain); curvature, computed as the second derivative of the surface elevation, useful to distinguish between convex, e.g., ridge, and concave, e.g., channel, surfaces; and flow, computed as the degree of flow accumulated on each cell (high values are indicative of stream or river channels).

\subsection{Experimental Settings}
%---------------------------------------------------------------------
\label{ssc:preparation_experiments}

To estimate the prior $P_s(\theta)$, we create $k=100$ bootstrap samples of the source dataset $T_s$ under uniform random sampling without replacement. We then record the $\theta^*$ that yields highest accuracy on each sample. In our experiments, $\theta$ corresponds to the number of hidden nodes in a neural network, $\theta \in [2, 50]$. For active learning, we divide dataset $T_t$ randomly into two (equal) parts: a pool of target instances from which data can be queried $T_t^1$, and a pool of test instances that remains unknown during training $T_t^2$. We then randomly generate 10 pairs of training and testing pools. We first limit our active-learning budget to $b=100$ queries, and subsequently study how accuracy varies with different budgets.

Regarding the posterior, we calculate the value of $\theta^*$ that maximizes the product of prior and likelihood. We then use $\theta^*$ to build an optimal model $f(\mathbf{x}|\theta^*)$ on the target pool $T_t^1$, and subsequently test it on the test pool $T_t^2$. In both domains, we have perfect knowledge of class labels (both source and target samples); class labels are hidden on the target sample to validate model performance.

Our hardware is made of a 3712-core computer cluster with 8 Tesla C2075 GPUs, and 22 GTK570 GPUs, 120TB Lustre Filesystem, and 127TB storage space.

\subsection{Methods for Comparison}
%----------------------------------
\label{ssc:other-methods}

For comparison purposes, our experiments include other domain adaptation techniques. We list and describe such techniques next.

\vspace*{2mm}
\noindent
Subspace Alignment \cite{fernando2013unsupervised}. The goal is to find separate subspaces for source and target domains using Principal Component Analysis, followed by a linear transformation that maps the source domain into the target domain . The result is an alignment of both spaces through the basis vectors. The number of principal components is optimized based on a theoretical bound.

\vspace*{2mm}
\noindent
Joint Distribution Optimal Transportation (JDOT) for Domain Adaptation  \cite{Courty17}.
The technique assumes a map that aligns the joint distributions ($\mathcal{X} \times \mathcal{Y}$) of the source and target domains. The optimization function combines both the distance between samples and the discrepancy in the loss between class labels.

\vspace*{2mm}
\noindent
Adaptation Regularization based Transfer Learning (ARTL) \cite{Long14b}. The central idea is to combine different strategies for transfer learning within a single framework: it simultaneously optimizes the structural risk functional over the source domain, the joint distribution matching of both marginal and conditional distributions, and the consistency of the geometric manifold corresponding to the marginal distribution.

\vspace*{2mm}
\noindent
Transfer Joint Matching (TJM) \cite{Long14}. The technique combines a shared representation between source and target domains with the concept of instance re-weighting, where source instances that fall within high density regions of the target domain, see their weight increased.

\vspace*{2mm}
\noindent
Transfer feature Learning with Joint Distribution Adaptation (JDA) \cite{Long13}. The technique jointly adapts both marginal and class-conditional probabilities using Principal Component Analysis.

\vspace*{2mm}
\noindent
Domain-Adversarial Training of Neural Networks (DATNN) \cite{Ganin16}. This is a neural network framework that implements domain adaptation by finding features that provide low error on the training set, while the features remain invariant across the two domains (i.e., across source and target domains). The architecture combines two learners that play in an adversarial manner: while one adjusts parameters to reduce training error, the other one adjusts parameters to discriminate (increase error) between source and target examples. The result is a regularized deep neural network that generates an informative abstract feature representation.

\vspace*{2mm}
\noindent
Geodesic Flow Kernel (GFK) for Unsupervised Domain Adaptation \cite{Grauman12}. This technique integrates an infinite number of subspaces between source and target domains by paying attention to geometric and statistical properties of both domains. The technique focuses on those subspaces that are domain invariant.

\subsection{Results}
%---------------------------------------------------------------------
\label{ssc:results}

\noindent
\textbf{Prior}. After estimating the sampling distribution for the optimal parameter $\theta^*$ on the source domain, our experiments show the following results. For Supernova: $\mu=33.75$ and $\sigma = 9.35$, and for Mars landforms: $\mu = 23.19$ and $\sigma = 12.63$. The range of values for the prior are set to $\theta^* \in [24, 43]$ for Supernova and $\theta^* \in [10, 36]$ for Mars landforms.

\begin{figure}[hbt]
\vspace*{0mm}
\begin{center}
\includegraphics[scale=0.22]{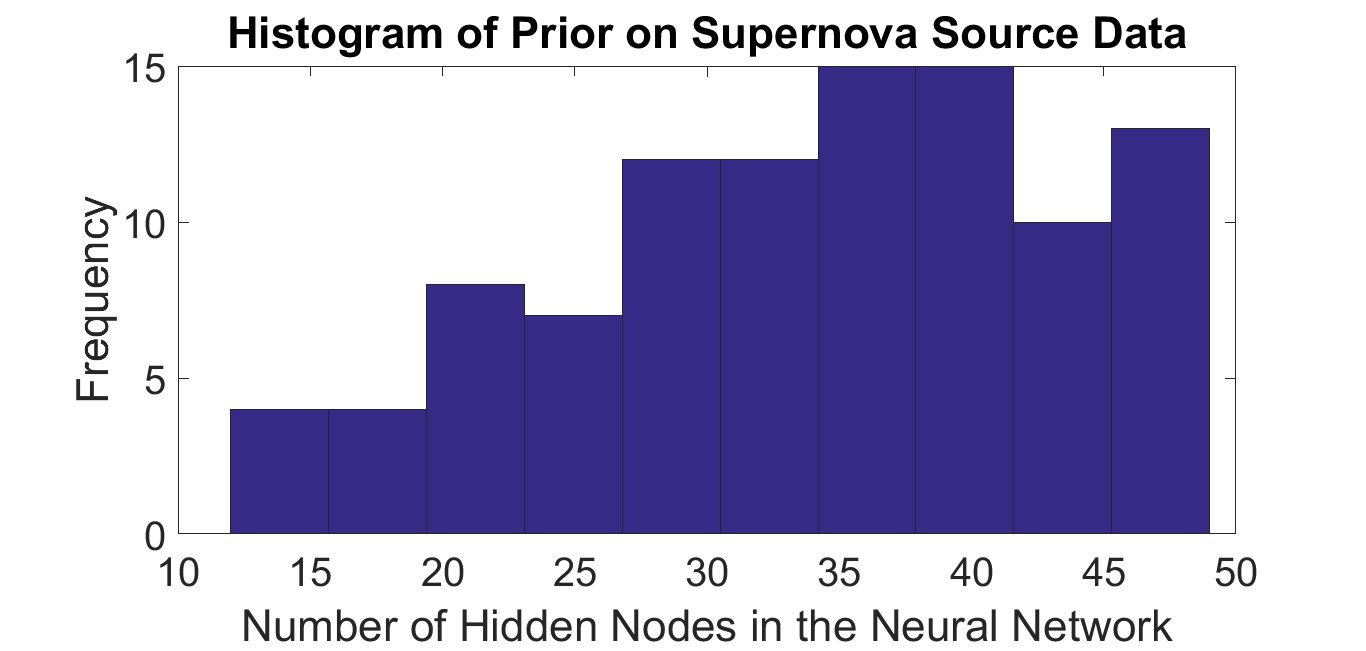}
\vspace*{0mm}
\caption{Histogram of optimal values of model complexity for the Supernova domain.}
\label{fig:supernova-histogram}
\end{center}
\vspace*{0mm}
\end{figure}

\begin{figure}[hbt]
\vspace*{0mm}
\begin{center}
\includegraphics[scale=0.22]{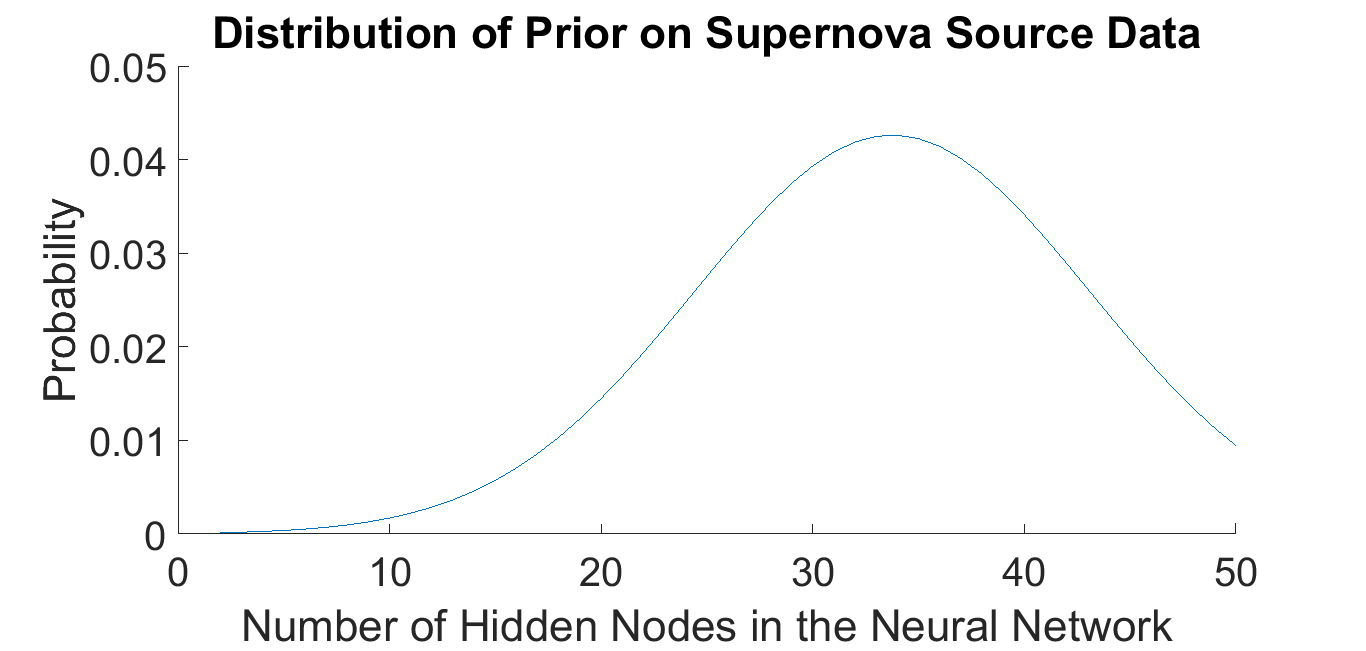}
\vspace*{0mm}
\caption{Gaussian approximation to the (prior) distribution of optimal model complexity values for the Supernova domain. }
\label{fig:supernova-approx}
\end{center}
\vspace*{-5mm}
\end{figure}

As an illustration, Figures~\ref{fig:supernova-histogram} and~\ref{fig:supernova-approx} show the histogram and corresponding univariate Gaussian approximation of optimal values of model complexity for the Supernova domain. Model complexity is measured in terms of the number of hidden nodes in a neural network. Approximating the prior distribution helps to narrow the search of values for the posterior, yielding substantial savings in computational cost (as shown in the following section).

\vspace*{5mm}

\vspace*{0mm}

\vspace*{0mm}
\begin{table*}[tbh]
\caption{Classification Accuracy (numbers enclosed in parentheses represent standard deviations). }
\vspace*{-5mm}
\hspace*{0mm}
\begin{center}
\begin{tabular}{|l||l|c|c|}
\hline
\multicolumn{1}{|c||}{General Method}               &
\multicolumn{1}{|c|}{Learning technique}            &
\multicolumn{2}{|c|}{Datasets}                      \\
\cline{3-4}
\multicolumn{1}{|c||}{}                             &
\multicolumn{1}{|c|}{}                              &
\multicolumn{1}{|c|}{Supernova Ia} 					&
\multicolumn{1}{|c|}{Mars Landforms}                \\
\hline
 & \textrm{Source Model}				& $\ 69.13\ (0.00) $ & $\ 74.36\ (9.40) $ \\
\cline{2-4}
 & \textrm{Subspace Alignment}         & $\ 62.56\ (7.98) $ & $\ 85.16\ (2.65) $ \\
\cline{2-4}
 & \textrm{JDOT SVM}                   & $\ 77.57\ (0.13) $ & $\ 85.2\ (0.59) $ \\
\cline{2-4}
\textrm{Domain} & \textrm{JDOT NN}     & $\ 69.05\ (0.08) $ & $\ 80.96\ (0.06) $ \\
\cline{2-4}
\textrm{Adaptation} & \textrm{DANN}    & $\ 80.4\ (0.3) $ & $\ 88.61\ (0.22) $ \\
\cline{2-4}
 & \textrm{TJM}                        & $\ 65.56\ ( 0.01) $ & $\ 82.28\ (0.03) $ \\
\cline{2-4}
 & \textrm{JDA}                        & $\ 70.64\ (0.03) $ & $\ 80.40\ (0.03) $ \\
\cline{2-4}
 & \textrm{ARTL}                       & $\ 66.21\ (0.01) $ & $\ 88.12\ (0.02) $ \\
\cline{2-4}
 & \textrm{GFK}                        & $\ 63.98\ (0.02) $ & $\  83.56\ (0.04) $ \\
\hline
\textrm{Source Model +} & \textrm{NN + AL}     & $\ 85.75\ (0.04) $ & $\ 80.41\ (0.08) $ \\
\cline{2-4}
\textrm{Active Learning} & \textrm{SVM + AL}  & $\ 69.33\ (0.17) $ & $\ 85.90\ (0.03) $ \\
\cline{2-4}
 & \textrm{LR + AL}   				   & $\ 83.70\ (0.03) $ & $\ 85.18\ (0.02) $ \\
\hline
\hline
\textbf{Bayesian DA} & \textbf{NN-DA-AL} & $\ \mathbf{86.17\ (0.35)} $ & $\  \mathbf{90.81\ (1.49)} $ \\
\hline
\end{tabular}
\end{center}
\label{tb:accuracy}
\vspace*{-3mm}
\end{table*}

\noindent
\textbf{Accuracy}. Table~\ref{tb:accuracy} shows average accuracy comparing our approach with two blocks of techniques: one block labeled ``Domain Adaptation" corresponding to the techniques described in Section~\ref{ssc:other-methods} (except for the first technique labeled ``Source Model" that simply builds a model on the source domain and applies it directly to the target domain).  The other block labeled ``Source Model + Active Learning" contains results for methods using active learning, with the initial model built on the source dataset. Our technique (Bayesian DA or NN-DA-AL) is shown as the last row of Table~\ref{tb:accuracy}. It combines domain adaptation with active learning using a neural network architecture (budget $b=100$ queries and the initial labeled pool is of size $r=10$).

For the first block, results show a significant increase in accuracy with our approach. For a statistical test, we use the Welch's Student Paired t-test distribution at the $p=0.01$ level. We also perform a multiple-comparison test by adjusting the statistical test using a Bonferroni adjustment \cite{Jensen00}; results are significant too at the $p=0.01$ level after the adjustment; this is true on both astronomical problems. These results show that using a posterior distribution of model complexity yields better classification accuracy on the target dataset than using the best prior. Results also show a major limitation of domain adaptation techniques founded on the assumption of the existence of feature-invariant subspaces between source and target domains. Real-world applications either do not guarantee the existence of such subspaces, or exhibit a complex subspace landscape where finding common subspaces turns difficult. Under the general assumption where both marginal and posterior probabilities between source and target domains differ ($P_{\rm s}(\mathbf{x}) \neq P_{\rm t}(\mathbf{x})$ and $P_{\rm s}(y | \mathbf{x}) \neq P_{\rm t}(y | \mathbf{x})$), a better strategy is to sample directly from the target domain under a framework that limits the cost of class queries. If the posterior class probability on the target domain follows a smooth distribution, a limited number of queries should suffice to attain an accurate predictive model.

The second block shows average accuracy with techniques that use active learning, where the initial model is built on the source domain. We report on a neural network (NN) with logistic activation units and 25 hidden nodes, Support  Vector Machines (SVM) with a radial basis function kernel, and Logistic Regression (LR); all with budget $b=100$ and $r=10$. We applied the same statistical test using Welch's Student Paired t-test distribution at the $p=0.01$ and a Bonferroni adjustment.  Our approach is significantly more accurate in all cases. This is true even with the use of a plain neural network, where there is no search for optimal complexity parameters (e.g., number of hidden units) and no source task to guide such search. Our method is also more efficient, as finding the best model-complexity from scratch on the target domain requires a new exhaustive search for an optimal value of model-complexity.

\begin{figure}[hbt]
\vspace*{0mm}
\begin{center}
\includegraphics[scale=0.35]{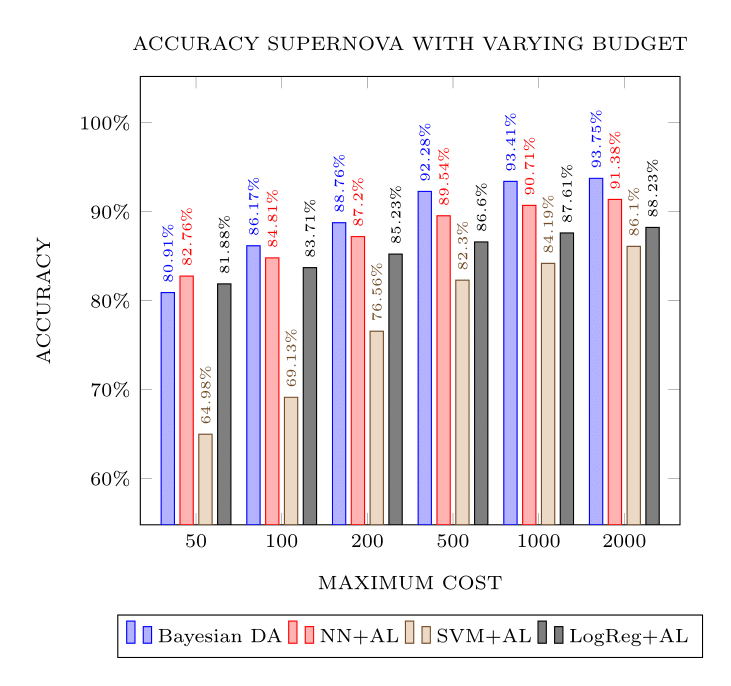}
\vspace*{-5mm}
\caption{Accuracy on Supernova improves significantly with increasing budget, and tends to converge after about 2,000 queries.}
\label{fig:AccSN-AL}
\end{center}
\vspace*{-3mm}
\end{figure}

\begin{figure}[hbt]
\vspace*{0mm}
\begin{center}
\includegraphics[scale=0.35]{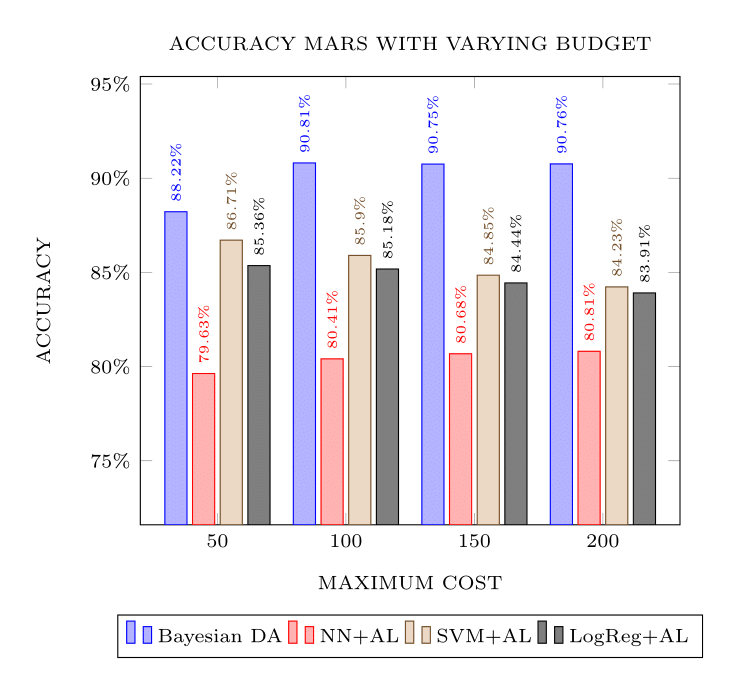}
\vspace*{-5mm}
\caption{Accuracy on Mars Landforms improves significantly with a budget of about 100 queries, and tends to converge after that threshold.}
\label{fig:AccMars-AL}
\vspace*{-3mm}
\end{center}
\end{figure}

The next experiment tests the impact of active learning within our approach as the budget is increased. Figure~\ref{fig:AccSN-AL} shows results for the Supernova task. Figure~\ref{fig:AccMars-AL} shows results for the Mars landforms task. For the Supernova task, there is a significant increase in accuracy as the budget grows from $50$ to $2,000$. This is expected as a large labeled sample on the target set provides enough evidence to generate accurate predictive models. Results tend to converge between $500-1000$ instances. For the Mars Landforms task, results tend to converge after only $100$ instances. In practical real-world scenarios, such results can be used to set a trade-off between the size of the budget and the cost of labeling new target instances. In the Supernova domain, for example, labeling new instances is extremely expensive as it involves running a full spectroscopic analysis; in such case, a lower budget may be preferred at the cost of some accuracy loss. The opposite is true on the Mars domain, where the cost of labeling segments with their correct landforms is relatively cheap, thus allowing for a budget increase.

\vspace*{0mm}

\vspace*{0mm}

\vspace*{1mm}
\noindent
\textbf{Execution Time}. A final experiment assesses the benefits gained in computational complexity between our approach and the common approach that finds a common subspace to match source and target distributions. Experiments follow the assumption that domain adaptation generates a prior distribution in the past, and hence does not incur any additional time during the current target task; in addition the search for an optimal value of model complexity on the target is limited to one standard deviation around the optimal value found on the source prior.  The model built on the source domain incurs on additional computational cost by searching for a common space over the two domains. We invoke Subspace Alignment as a representative case of feature matching techniques. For the Supernova task, execution time is reduced from about $90$ hrs to under $2$ hours. For the Mars landforms task, execution time goes from about $4$ hrs to about $4$ minutes. Results show the advantage of generating a prior distribution of model complexity on the source domain that is readily available on a new target task: it obviates an exhaustive search for an optimal parameter value.

\section{Summary and Conclusions}
%********************************
\label{sc:summary}

We propose a new direction in domain-adaptation using a Maximum a Posteriori approach where the prior distribution is obtained from a source task (previous experience), whereas the likelihood is obtained from the target (or current) task. Our methodology invokes active learning to compensate for the lack of (target) class labels, leaving the budget size as an experimental parameter. Our study leads to a new formulation of the likelihood as a function of empirical error and a term that depends on model complexity as estimated by the Vapnik-Chervonenkis dimension. Overall, our technique broadens the general applicability of domain adaptation by relaxing the stringent requirement of close proximity between source and target distributions.

Empirical results on two astronomical problems show a significant advantage in computational cost as the range of complexity values on the target domain is limited to a small window; this is the result of using a prior distribution over the complexity parameter derived from the source domain. In terms of accuracy, results show a significant increase in performance with our approach; this holds for both astronomical domains. Our experiments also show  a trade-off between budget size and the cost of labeling; in cases where labeling is relatively cheap, one can increase the budget to achieve an increase in accuracy performance.

As future work, we will investigate how to extend our work when multiple source domains are available. One possibility is to simply choose the best prior based on domain knowledge, or through a ranking system that orders all source tasks based on spatial or temporal proximity to the astronomical event of interest. Another direction is to combine all priors by assigning a degree of relevance to each source task. The posterior distribution can then be defined as a weighted combination of all available priors.

\iffalse
We also plan to investigate how to extend the applicability of domain adaptation when combined with active learning. In particular, we are interested to devise an efficient strategy to sample examples from the target domain. The existence of a source dataset can help optimize such sampling strategy by first capturing the complexity of the source task, and then using such measure to sample on the target. We plan to devise new task-complexity measures to guide several sampling strategies. For example, complex tasks are those where classes alternate abruptly throughout local regions of the input space \cite{Rendell90}; here an efficient sample strategy ought to cover a broad portion of the input space to account for the expected lack of class uniformity; focusing on a small region of the input space is prone to miss other similarly complex regions. On the other hand, easy tasks are characterized by large regions where class labels remain constant, and few peaks characterize the overall input-output landscape \cite{Rendell90}; here a sample strategy can focus on narrow regions of the input space where simple decision boundaries suffice to attain good generalization performance.
\fi

Additionally we hope to stimulate the astronomical community to consider domain adaptation as a useful resource when analyzing different surveys on similar objects. For example, while providing class labels for transient objects or events contained in one single survey is still feasible --even though costly-- the ability of labeling variable sources across the large number of available surveys is almost non-existent. The goal of acquiring predictive models from many surveys is a daunting task. This can be tackled by creating predictive models that adapt across datasets under analysis using domain adaptation techniques. The need for domain adaptation lies in the distributional discrepancy between source and target domains.

%========================================================================

\section*{Acknowledgments}

This work was partly supported by the Center for Advanced Computing and Data Systems (CACDS), and by the Texas Institute for Measurement, Evaluation, and Statistics (TIMES) at the University of Houston.

%========================================================================
% Bibliography

\end{document}